%% file: iclr2023_conference.tex
\documentclass{article} 
\usepackage{iclr2023_conference,times}

\iclrfinalcopy

\input{math_commands.tex}

\usepackage{hyperref}
\usepackage{url}

\usepackage{amsmath}
\usepackage{booktabs}
\usepackage{subcaption}
\usepackage{graphicx}
\usepackage[utf8]{inputenc}
\usepackage{hyperref}
\usepackage{mathtools}
\usepackage{amsfonts}
\usepackage{multirow}
\usepackage{wasysym}
\usepackage{adjustbox}
\usepackage{array}
\usepackage{xfrac}

\usepackage{algorithm}
\usepackage[noend]{algpseudocode}

\usepackage{pifont}

\title{\scalebox{1.0}[1.0]{(Certified!!) Adversarial Robustness for Free!}}

\author{Nicholas Carlini\thanks{Joint first authors}$^{*1}$ \quad Florian Tram\`er$^{*1}$ \quad Krishnamurthy (Dj) Dvijotham$^1$ \\ \textbf{Leslie Rice}$^2$ \quad \textbf{Mingjie Sun}$^2$ \quad \textbf{J. Zico Kolter}$^{2,3}$ \\ $^1${Google} \qquad $^2${Carnegie Mellon University} \qquad $^3${Bosch Center for AI}}

\date{}
\newcommand{\cmark}{\ding{51}}%
\newcommand{\xmark}{\ding{55}}%
\newcommand{\br}[1]{\left({#1}\right)}
\newcommand{\N}{\mathcal{N}}
\usepackage{xcolor}
\newcommand{\zico}[1]{\textcolor{red}{[Zico: #1]}}

\begin{document}

            \vspace{-1em}
\maketitle

\vspace{-1em}
\begin{abstract}
    In this paper we show how to achieve state-of-the-art
    \emph{certified} adversarial robustness to $\ell_2$-norm bounded perturbations by relying exclusively on off-the-shelf 
    pretrained models.
    To do so, we instantiate the denoised smoothing approach of \citet{salman2020denoised}~by combining a
    pretrained denoising diffusion probabilistic model and a standard high-accuracy classifier.
    This allows us to certify 71\% accuracy on ImageNet under adversarial perturbations constrained to be within an $\ell_2$ norm of $\varepsilon=0.5$, an improvement of 
    14 percentage points over the prior certified SoTA using any approach,
    or an improvement of 30 percentage points over denoised smoothing.
    %
    We obtain these results using only pretrained diffusion models and image classifiers, without requiring any fine tuning or retraining of model parameters.
\end{abstract}

\section{Introduction}
Evaluating the robustness of deep learning models to norm bounded adversarial perturbations has been shown to be difficult \citep{athalye2018obfuscated, uesato2018adversarial}. Certified defenses---such as those based on bound propagation \citep{gowal2018effectiveness, mirman2018differentiable} or randomized smoothing \citep{lecuyer2019certified, cohen2019certified}---offer provable guarantees that a model's predictions are robust to norm-bounded adversarial perturbations, for a large fraction of examples in the test set.



The current state-of-the-art approaches to certify robustness to adversarial perturbations bounded in the $\ell_2$ norm rely on \emph{randomized smoothing} \citep{lecuyer2019certified,cohen2019certified}. 
By taking a majority vote over the labels predicted by a ``base classifier'' under random Gaussian perturbations of the input, if the correct class is output sufficiently often, then the defense's output on the original un-noised input is guaranteed to be robust to $\ell_2$ norm bounded adversarial perturbations.

Denoised smoothing \citep{salman2020denoised} is a certified defense that splits this one-step process into two.
After randomly perturbing an input,
the defense first applies a \emph{denoiser} model that aims to remove the added noise, followed by a standard
\emph{classifier} that guesses a label given this \emph{noised-then-denoised} input. 
This enables applying randomized smoothing to pretrained black-box base classifiers, as long as the denoiser can produce clean images close to the base classifier's original training distribution.
%

We observe that the recent line of work on
\emph{denoising diffusion probabilistic models} \citep{sohl2015deep,ho2020denoising,nichol2021improved}---which achieve
state-of-the-art results on image generation---are a perfect match for the denoising step in a denoised smoothing defense.
%
A forward diffusion process takes a source data distribution (e.g., images from some data distribution)
and then adds Gaussian noise until the distribution converges to a high-variance isotropic Gaussian.
Denoising diffusion models are trained to invert this process.
Thus, we can use a diffusion model as a denoiser that recovers high quality denoised inputs from inputs perturbed with Gaussian noise.  

In this paper, we combine state-of-the-art, publicly available diffusion models as denoisers with standard pretrained state-of-the-art classifiers.
We show that the resulting denoised smoothing defense obtains significantly \emph{better} certified robustness results---for perturbations of $\ell_2$ norm of $\epsilon \leq 2$ on ImageNet and $\epsilon \leq 0.5$ on CIFAR-10---compared to the ``custom'' denoisers trained in prior work~\citep{salman2020denoised}, or in fact with any certifiably robust defense (even those that do not rely on denoised smoothing). Code to reproduce our experiments is available at: \url{https://github.com/ethz-privsec/diffusion_denoised_smoothing}.

\section{Background}

\paragraph{Adversarial examples} \citep{biggio2013evasion,szegedy2013intriguing} are inputs $x'=x+\delta$ constructed by taking
some input $x$ (with true label $y \in \mathcal{Y}$) and adding a perturbation $\delta$ (that is assumed to be imperceptible and hence label-preserving) so that a given classifier $f$ misclassifies the perturbed input, i.e., $f(x + \delta) \neq y$.
The ``smallness'' of $\delta$ is quantified by its Euclidean norm, and we constrain
$\lVert \delta \rVert_2 \leq \varepsilon$.
Even when considering exceptionally small perturbation budgets
(e.g., $\varepsilon=0.5$) modern classifiers often have near-0\%
accuracy \citep{carlini2017towards}.

\paragraph{Randomized smoothing} \citep{lecuyer2019certified, cohen2019certified}
is a technique to certify the robustness of arbitrary classifiers against
adversarial examples under the $\ell_2$ norm.
Given an input $x$ and base classifier $f$, randomized smoothing considers a smooth version of $f$ defined as:
\begin{align}
    g(x) \coloneqq \text{argmax}_{c} \Pr_{\delta \sim \mathcal{N}\br{0, \sigma^2 \mathbf{I}}}(f(x + \delta) = c) \label{eq:rs}
\end{align}

\citet{cohen2019certified} prove that the smooth classifier $g$ is robust to perturbations of $\ell_2$ radius $R$, where the radius $R$ grows with the classifier's ``margin'' (i.e., the difference in probabilities assigned to the most likely and second most-likely classes).
As the probability in \Eqref{eq:rs} cannot be efficiently computed when the base classifier $f$ is a neural network, \citet{cohen2019certified}~instantiate this defense by sampling a small number $m$ of noise instances (e.g., $m=10$) and taking a majority vote over the outputs of the base classifier $f$ on $m$ noisy versions of the input. To compute a lower-bound on this defense's robust radius $R$, they estimate the probabilities $\Pr[f(x + \delta) = c]$ for each class label $c$ by sampling a large number $N$ of noise instances $\delta$ (e.g., $N=100{,}000$). See~\citet{cohen2019certified} for details.

\paragraph{Denoised smoothing} 
\citep{salman2020denoised} is an instantiation of randomized smoothing, where the base classifier $f$ is composed of a denoiser $\texttt{denoise}$ followed by a standard classifier $f_\texttt{clf}$:
\begin{equation}
    \label{eq:ds}
    f(x+\delta) \coloneqq f_{\texttt{clf}}(\texttt{denoise}(x + \delta)) \;.
\end{equation}

Given a very good denoiser (i.e., $\texttt{denoise}(x + \delta) \approx x$ with high probability for $\delta \sim \N\br{0, \sigma^2 \mathbf{I}}$), we can expect the base classifier's accuracy on noisy images to be similar to the \emph{clean} accuracy of the standard classifier $f_{\texttt{clf}}$.
\citet{salman2020denoised} instantiate their denoised smoothing technique by training custom denoiser models with Gaussian noise augmentation, combined with off-the-shelf pretrained classifiers.

\paragraph{Denoising Diffusion Probabilistic Models} \citep{sohl2015deep,ho2020denoising,nichol2021improved} are a form of generative
model that work by learning a model that can reverse time on a diffusion process of the form $x_t \sim \sqrt{1-\beta_t} \cdot x_{t-1} + \beta_t \cdot \omega_t, \omega_t \sim \N\br{0, \mathbf{I}}$ with $x_0$ coming from the data distribution, and the $\beta_t$ being fixed (or learned) variance parameters. The diffusion process transforms images from the target data distribution to purely random noise over time. The reverse process then synthesizes images from the data distribution starting with random Gaussian noise. In this paper we will not make use of diffusion models in the typical way;
instead it suffices to understand just one single property about how they are
trained.

Given a clean training image $x \in [-1,1]^{w \cdot h \cdot c}$, 
a diffusion model selects a \emph{timestep} $t \in \mathbb{N}^+$ from some fixed schedule and then samples a noisy image $x_t$ of the form
\begin{align}
\label{eq:q_sample}
    x_t &\coloneqq \sqrt{\alpha_t} \cdot x + \sqrt{1-\alpha_t} \cdot \mathcal{N}(0, \mathbf{I}) \;,
\end{align}
where the factor $\alpha_t$ is a constant derived from the timestamp $t$ that determines the amount of noise to be added to the image (the noise magnitude increases monotonically with $t$).

The diffusion model is then trained (loosely speaking) to minimize
the discrepancy between $x$ and $\texttt{denoise}(x_t; t)$;
that is, to predict what the original (un-noised) image should look like
after applying the noising step at timestep $t$.\footnote{State-of-the-art diffusion models are actually trained to predict the \emph{noise} rather than the denoised image directly~\citep{ho2020denoising, nichol2021improved}.}

\section{Diffusion Denoised Smoothing}
Our approach, Diffusion Denoised Smoothing (DDS), requires no new technical ideas
on top of what was introduced in the section above.

\begin{figure}
\algrenewcommand\algorithmicprocedure{}

\begin{minipage}[t]{0.39\textwidth}
\begin{algorithm}[H]
 \caption{Noise, denoise, classify}
 \label{alg}
 \begin{algorithmic}[1]
    \Procedure{\texttt{NoiseAndClassify}}{$x, \sigma$}:
    \State $t^\star, \alpha_{t^\star} \gets \Call{\texttt{GetTimestep}}{\sigma}$
    \State $x_{t^\star} \gets \sqrt{\alpha_{t^\star}} (x + \mathcal{N}(0, \sigma^2 \mathbf{I}))$
    \State $\hat{x} \gets \texttt{denoise}(x_{t^\star}; t^\star)$  
    \State $y \gets f_{\textrm{clf}}(\hat{x})$ 
    \State \Return $y$
    \EndProcedure
    \State
    \Procedure{\texttt{GetTimestep}}{$\sigma$}:
    \State $t^\star \gets \textrm{find } t \,\,\textrm{s.t.}\ {1- \alpha_t \over \alpha_t} = \sigma^2$
    \State \Return $t^\star, \alpha_{t^\star}$
  \EndProcedure
 \end{algorithmic}
\end{algorithm}
\end{minipage}
\hfill
\begin{minipage}[t]{0.58\textwidth}
\begin{algorithm}[H]
 \caption{Randomized smoothing~\citep{cohen2019certified}}
 \label{alg:rs}
 \begin{algorithmic}[1]
 \Procedure{\texttt{Predict}}{$x, \sigma, N, \eta$}:
    \State $\texttt{counts} \gets \mathbf{0}$
    \For{$i \in \{1, 2, \ldots, N\}$}
    \State $y \gets \Call{\texttt{NoiseAndClassify}}{x, \sigma}$
    \State $\texttt{counts}[y] \gets \texttt{counts}[y] + 1$
    \EndFor
    \State $\hat{y}_A, \hat{y}_B \gets \textrm{top two labels in \texttt{counts}}$
    \State $n_A, n_B \gets \texttt{counts}[\hat{y}_A], \texttt{counts}[\hat{y}_B]$
    \If{$\Call{\texttt{BinomPTest}}{n_A, n_A + n_B, \sfrac{1}{2}} \leq \eta$}
      \State \Return $\hat{y}_A$
      \Else
      \State \Return $\texttt{Abstain}$
  \EndIf
 \EndProcedure
 \end{algorithmic}
\end{algorithm}
\end{minipage}

 \caption{Our approach can be implemented in under 15 lines of code, given an off-the-shelf classifier $f_{\textrm{clf}}$ and an off-the-shelf diffusion model \texttt{denoise}. The \textsc{Predict} function is adapted from \citet{cohen2019certified} and takes as input a number of noise samples $N$ and a statistical significance level $\eta \in (0, 1)$ and inherits the same robustness certificate proved in \citet{cohen2019certified}.}
 \label{fig:main}
\end{figure}

\paragraph{Denoised smoothing via a diffusion model.} The only minor technicality required for our method is to map between the noise model required by randomized smoothing and the noise model used within diffusion models.  Specifically, randomized smoothing requires a data point augmented with additive Gaussian noise
$    x_{\mathrm{rs}} \sim \mathcal{N}(x, \sigma^2 \mathbf{I})$,
whereas diffusion models assume the noise model
$x_t \sim \mathcal{N}(\sqrt{\alpha_t} x, (1-\alpha_t) \mathbf{I})$.
Scaling $x_{\mathrm{rs}}$ by $\sqrt{\alpha_t}$ and equating the variances yields the relationship
\begin{equation}\label{eq:solve_sigma}
\sigma^2 = \frac{1 - \alpha_t}{\alpha_t} \;.
\end{equation}
Thus, in order to employ a diffusion model for randomized smoothing at a given noise level $\sigma$, we first find the timestep $t^\star$ such that $\sigma^2 = \frac{1 - \alpha_{t^\star}}{\alpha_{t^\star}}$; the precise formula for this equation will depend on the schedule of the $\alpha_t$ terms used by the diffusion model, but this can typically be computed in closed form, even for reasonably complex diffusion schedules.\footnote{For example, in \citet{nichol2021improved}, the authors advocate for the schedule
$    \alpha_t = f(t)/f(0), \mbox{ where } f(t) = \cos \left (\frac{t/T + s}{1 + s}\cdot \frac{\pi}{2} \right)^2$
for various values of $T$, and $s$ discussed in this reference. In this case, for a given desired value of $\sigma^2$, some algebra yields the solution for $t$
\begin{equation}
    t^\star = T\left (1 - \frac{2(1+s)\csc^{-1}\left (\sqrt{1+\sigma^2} \csc \left(\frac{\pi}{2+2s}\right) \right )}{\pi} \right ).
\end{equation}
The actual formula here is unimportant and only shown as an illustration of how such computation can look in practice.  Even when such a closed form solution does not exist, because the schedules for $\alpha_t$ are monotonic decreasing, one can always find a solution via 1D root-finding methods if necessary.} Next, we compute
\begin{equation}
    x_{t^\star} = \sqrt{\alpha_{t^\star}} (x + \delta), \;\; \delta \sim \mathcal{N}(0,\sigma^2 \mathbf{I})
\end{equation}
and apply the diffusion denoiser on $x_{t^\star}$ to obtain an estimate of the denoised sample
\begin{equation}
    \hat{x} =  \texttt{denoise}(x_{t^\star};t^\star) \;.
\end{equation}
And finally, we classify the estimated denoised image with an off-the-shelf classifier
\begin{equation}
    y = f_{\text{clf}}(\hat{x}) \;.
\end{equation}
The entirety of this algorithmic approach is shown in Figure~\ref{fig:main}.

To obtain a robustness certificate, we repeat the above denoising process many times
(e.g., 100{,}000) and compute the certification radius using the approach of \citet{cohen2019certified} (note that since our diffusion models expects inputs in $[-1, 1]^d$, we then divide the certified radius by $2$ to obtain a certified radius for inputs in $[0,1]$ as assumed in all prior work).


\paragraph{One-shot denoising.}
Readers familiar with diffusion models may recall that
the standard process repeatedly applies a ``single-step'' denoising operation $x_{t-1}=d(x_t; t)$ that aims to convert a noisy image at some timestep $t$ to a (slightly less) noisy image at the previous timestep $t-1$. The full diffusion process would then be defined by the following iterative procedure:
\[
\tilde{x} = \texttt{denoise}_{\text{iter}}(x + \delta; t) \coloneqq d(d(\dots d(d(x + \delta; t); t-1)\dots; 2); 1) \;.
\]
In fact, each application of the one-step denoiser $d$ consists of two steps: (1) an estimation of the fully denoised image $x$ from the current timestep $t$, and (2) computing a (properly weighted, according to the diffusion model) average between this estimated denoised image and the noisy image at the previous timestep $t-1$. Thus, instead of performing the entire $t$-step diffusion process to denoise an image, it is also possible to run the diffusion step $d$ \emph{once} and simply output the best estimate for the denoised image $x$ in one shot.


When a diffusion model generates images from scratch (i.e., the denoiser is applied to pure noise), the iterative process gives higher fidelity outputs than this one-shot approach~\citep{ho2020denoising}.
But here, where we aim to denoise one particular image, a one-shot approach has two advantages:
\begin{enumerate}
    \item \textbf{High accuracy}: it turns out that standard pretrained classifiers are more accurate on
    one-shot denoised images compared to images denoised with the full $t$-steps of denoising.
    We hypothesize this is due to the fact that when we first apply the single-step denoiser $d$ at timestep $t$, the denoiser already has all the available information about $x$. By applying the denoiser multiple times, we can only \emph{destroy} information about $x$ as each step adds new (slightly smaller) Gaussian noise.
    In fact, by using the iterative $t$-step denoising strategy, we are in essence pushing part of the classification task onto the denoiser, in order to decide how to fill in the image. 
    Section~\ref{sec:analysis} experimentally validates this hypothesis.
    
    \item \textbf{Improved efficiency}: instead of requiring several hundred (or thousand)
    forward passes to denoise any given image, we only require one single pass.
    This is especially important when we perform many thousand predictions
    as is required for randomized smoothing to obtain a robustness certificate.
\end{enumerate}

\paragraph{Related work.}
We are not the first to observe a connection between randomized smoothing
and diffusion models. The work of \citet{leeprovable} first studied this problem---however they do not obtain significant accuracy improvements, likely due to the fact that diffusion models available at the time that work was done were not good enough.  Separately, \citet{nie2022diffusion} suggest that diffusion models might be able to
provide strong \emph{empirical} robustness to adversarial examples, as evaluated by robustness under adversarial attacks computed using existing attack algorithms;
this is orthogonal to our results.

\section{Evaluation}

We evaluate diffusion denoised smoothing on two standard datasets, CIFAR-10 and ImageNet,
and find it gives state-of-the-art certified $\ell_2$ robustness on both.
On CIFAR-10, we draw $N=100{,}000$ noise samples and on ImageNet we draw $N=10{,}000$ samples to certify the robustness following \citet{cohen2019certified}.

As is standard in prior work, we perform randomized smoothing for three different noise magnitudes, $\sigma \in \{0.25, 0.5, 1.0\}$. For a fair comparison to prior work in Table~\ref{tab:cifar10_sota} and Table~\ref{tab:imagenet_sota}, we give the best results reported in each paper across these same three noise magnitudes.
Note that prior work only uses three levels of noise due to the computational overhead;
one benefit of using a diffusion model is we could have used other amounts of noise
without training a new denoiser model.

\paragraph{CIFAR-10 configuration.}
We denoise CIFAR-10 images with the 50M-parameter diffusion model from~\citet{nichol2021improved}.\footnote{\url{https://github.com/openai/improved-diffusion}} The denoised images are classified with a 87M-parameter ViT-B/16 model~\citep{dosovitskiy2020image} that was pretrained on ImageNet-21k~\citep{deng2009imagenet} (in $224\times224$ resolution) and finetuned on CIFAR-10. We use the implementation from HuggingFace\footnote{\url{https://huggingface.co/aaraki/vit-base-patch16-224-in21k-finetuned-cifar10}} which reaches $97.9\%$ test accuracy on CIFAR-10.
In addition, we also report results with a standard 36M parameter Wide-ResNet-28-10 model~\citep{zagoruyko2016wide} trained on CIFAR-10 to 95.2\% accuracy.

As is typical, we report results with images normalized to $[0,1]^{32\times32\times3}$.
We obtain a throughput of $825$ images per second through the diffusion model and ViT classifier on an A100 GPU at a batch size of $1{,}000$. 
We report robust accuracy results averaged over the entire CIFAR-10 test set.

\paragraph{ImageNet configuration.}
We denoise ImageNet images with the 552M-parameter class-unconditional diffusion model from \citet{dhariwal2021diffusion},
and classify images with the 305M-parameter BEiT large model \citep{bao2021beit}
which reaches a $88.6\%$ top-1 validation accuracy
using the implementation from \texttt{timm} \citep{rw2019timm}.
We report results for our images when normalized to $[0,1]^{224\times224\times3}$ to allow us to compare to prior work.
The overall latency of this joint denoise-then-classify
model is $1.5$ seconds per image on an A100 GPU at a batch size of $32$.
We report results averaged over $1{,}000$ images randomly selected
from the ImageNet test set.

\begin{table}
    \centering
    \setlength{\tabcolsep}{5pt}
        \resizebox{1.0\textwidth}{!}{
    \begin{tabular}{@{}l c c r r r r r@{}}
        & & & \multicolumn{5}{c}{Certified Accuracy at $\varepsilon$ (\%)} \\
         \cmidrule{4-8}
        Method & Off-the-shelf & Extra data & $0.5$ & $1.0$ & $1.5$ & $2.0$ & $3.0$ \\
        \toprule
        PixelDP~\citep{lecuyer2019certified} & \Circle & \xmark & $^{(33.0)}$16.0  & -  & -\\ 
        RS~\citep{cohen2019certified} & \Circle & \xmark & $^{(67.0)}$49.0  & $^{(57.0)}$37.0  & $^{(57.0)}$29.0  & $^{(44.0)}$19.0 &  $^{(44.0)}$12.0\\
        SmoothAdv~\citep{salman2019provably} & \Circle & \xmark & $^{(65.0)}$56.0 & $^{(54.0)}$43.0 & $^{(54.0)}$37.0 & $^{(40.0)}$27.0 & $^{(40.0)}$20.0\\
        Consistency~\citep{jeong2020consistency} & \Circle & \xmark & $^{(55.0)}$50.0 & $^{(55.0)}$44.0 & $^{(55.0)}$34.0 & $^{(41.0)}$24.0 & $^{(41.0)}$17.0\\
        MACER~\citep{zhai2020macer} & \Circle & \xmark & $^{(68.0)}$57.0 & $^{(64.0)}$43.0 & $^{(64.0)}$31.0 & $^{(48.0)}$25.0 & $^{(48.0)}$14.0\\
        Boosting~\citep{horvath2021boosting} & \Circle & \xmark & $^{(65.6)}$57.0 & $^{(57.0)}$44.6 & $^{(57.0)}$\bf38.4 & $^{(44.6)}$\bf28.6 & $^{(38.6)}$\bf21.2\\
        DRT~\citep{yang2021certified} & \Circle & \xmark & $^{(52.2)}46.8$ & $^{(55.2)}44.4$ & \bf$^{(49.8)}$39.8 & \bf$^{(49.8)}$30.4 & $^{(49.8)}$\bf23.4 \\
        SmoothMix~\citep{jeong2021smoothmix} & \Circle & \xmark & $^{(55.0)}$50.0 & $^{(55.0)}$43.0 & $^{(55.0)}$\bf38.0 & $^{(40.0)}$26.0 & $^{(40.0)}$20.0\\
        ACES~\citep{horvath2022robust} & \LEFTcircle & \xmark & $^{(63.8)}$54.0 & $^{(57.2)}$42.2 & $^{(55.6)}$35.6 & $^{(39.8)}$25.6 & $^{(44.0)}$19.8\\
        \midrule
        Denoised~\citep{salman2020denoised} & \LEFTcircle & \xmark & $^{(60.0)}$33.0 & $^{(38.0)}$14.0 & $^{(38.0)}$6.0 & - & -\\
        Lee~\citep{leeprovable} & \CIRCLE & \xmark &  41.0 & 24.0 & 11.0 & - & -\\
        \midrule
        \textbf{Ours} &  \CIRCLE & \cmark &  $^{(82.8)}$\bf71.1 & $^{(77.1)}$\bf54.3 & $^{(77.1)}$\bf38.1 & $^{(60.0)}$\bf29.5 & $^{(60.0)}$13.1 \\
        \bottomrule
    \end{tabular}
    }
    \caption{ImageNet certified top-1 accuracy for prior defenses on randomized smoothing and denoised smoothing.
    Randomized smoothing techniques rely on special-purpose models (indicated by a empty circle). The work of \citet{horvath2022robust} is an exception in that it selectively applies either a robust or accurate off-the-shelf classifier (indicated by a half full circle).
    Denoised smoothing \citep{salman2020denoised} use an off-the-shelf classifier but train their own denoiser (indicated by a half full circle).
    Our base approach uses an off-the-shelf classifier and off-the-shelf denoiser (indicated by a full circle).
    Each entry lists the certified accuracy, with the clean accuracy for that model in parentheses, using numbers taken from respective papers.}
    \label{tab:imagenet_sota}
\end{table}

\begin{table}
    \centering
        \resizebox{1.0\textwidth}{!}{
    \begin{tabular}{@{}l c c r r r r@{}}
        & & & \multicolumn{4}{c}{Certified Accuracy at $\varepsilon$ (\%)} \\
         \cmidrule{4-7}
        Method & Off-the-shelf & Extra data & $0.25$ & $0.5$ & $0.75$ & $1.0$ \\
        \toprule
        PixelDP~\citep{lecuyer2019certified} & \Circle & \xmark & $^{(71.0)}$22.0 & $^{(44.0)}$2.0 & -  & - \\
        RS~\citep{cohen2019certified} & \Circle & \xmark & $^{(75.0)}$61.0 & $^{(75.0)}$43.0 & $^{(65.0)}$32.0  & $^{(66.0)}$22.0 \\
        SmoothAdv~\citep{salman2019provably} & \Circle & \xmark & $^{(75.6)}$67.4 & $^{(75.6)}$57.6 & $^{(74.8)}$47.8 & $^{(57.4)}$38.3 \\
        SmoothAdv~\citep{salman2019provably} & \Circle & \cmark & $^{(84.3)}$74.9 & $^{(80.1)}$63.4 & $^{(80.1)}$\bf51.9 & $^{(62.2)}$\bf39.6\\
        Consistency~\citep{jeong2020consistency} & \Circle & \xmark & $^{(77.8)}$68.8  & $^{(75.8)}$58.1 & $^{(72.9)}$48.5 & $^{(52.3)}$37.8  \\
        MACER~\citep{zhai2020macer} & \Circle & \xmark & $^{(81.0)}$71.0  & 
        $^{(81.0)}$59.0  & $^{(66.0)}$46.0 & $^{(66.0)}$38.0 \\
        Boosting~\citep{horvath2021boosting} & \Circle & \xmark & $^{(83.4)}$70.6  & $^{(76.8)}$60.4  & \bf $^{(71.6)}$52.4  & $^{(52.4)}$\bf38.8 
        \\
        DRT~\citep{yang2021certified} & \Circle & \xmark & $^{(81.5)}70.4$ & $^{(72.6)}60.2$ &  $^{(71.9)}$50.5 & \bf $^{(56.1)}$39.8\\
        SmoothMix~\citep{jeong2021smoothmix} & \Circle & \xmark & $^{(77.1)}$67.9  & $^{(77.1)}$57.9  & $^{(74.2)}$47.7  & $^{(61.8)}$37.2 \\
        ACES~\citep{horvath2022robust} & \LEFTcircle & \xmark & $^{(79.0)}$69.0 & $^{(74.2)}$57.2 & $^{(74.2)}$47.0 & $^{(58.6)}$37.8 \\
        \midrule
        Denoised~\citep{salman2020denoised} & \LEFTcircle & \xmark & $^{(72.0)}$56.0  & $^{(62.0)}$41.0  & $^{(62.0)}$28.0  & $^{(44.0)}$19.0 \\
        Lee~\citep{leeprovable} & \CIRCLE & \xmark & 60.0 & 42.0 & 28.0 & 19.0 \\
        \midrule
        \textbf{Ours} &  \CIRCLE & \cmark & $^{(88.1)}$76.7  & $^{(88.1)}$63.0  & $^{(88.1)}$45.3  & $^{(77.0)}$32.1\\
        \midrule
        \textbf{Ours (+finetuning)} & \LEFTcircle & \cmark & \bf $^{(91.2)}$79.3  & \bf $^{(91.2)}$65.5  & $^{(87.3)}$48.7  & $^{(81.5)}$35.5\\
        \bottomrule
    \end{tabular}
    }
    \caption{CIFAR-10 certified accuracy for prior defenses from the literature.
    The columns have the same meaning as in Table~\ref{tab:imagenet_sota}.}
    \label{tab:cifar10_sota}
\end{table}

\subsection{Results}
\begin{figure}
    \begin{subfigure}[b]{0.48\textwidth}
    \centering
        \includegraphics[width=\textwidth]{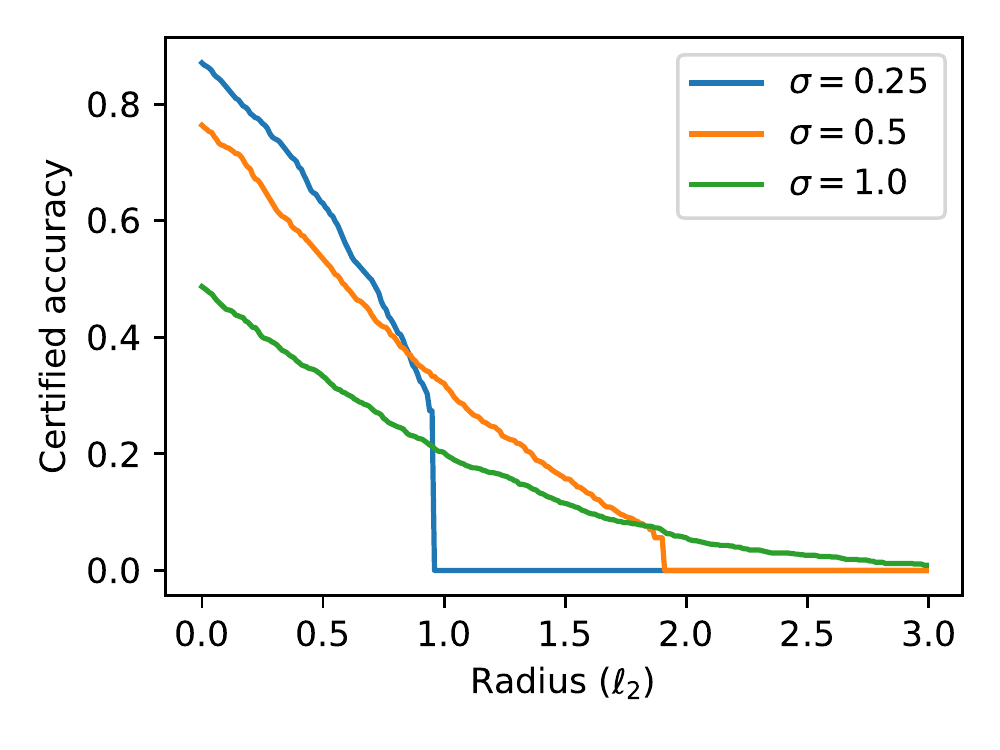}
        \vspace{-3ex}
        \label{fig:cifar10}
    \caption{CIFAR-10}
    \end{subfigure}
    \hfill
    \begin{subfigure}[b]{0.48\textwidth}
    \centering
        \includegraphics[width=\textwidth]{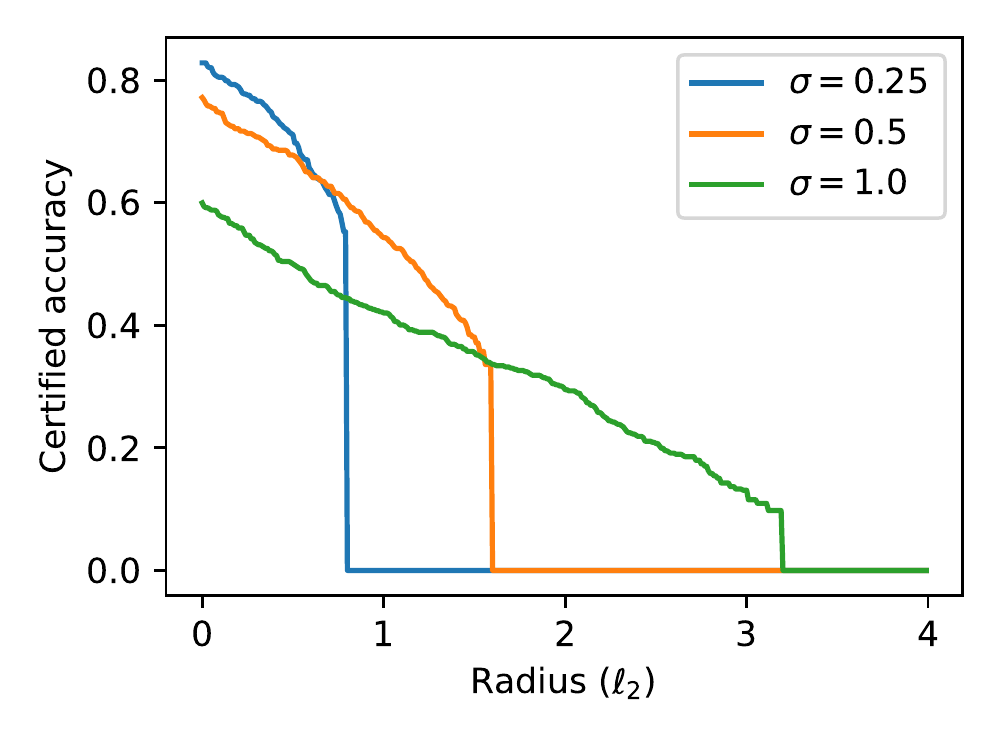}
        \vspace{-3ex}
        \label{fig:imagenet}
    \caption{ImageNet}
    \end{subfigure}
    \vspace{-1ex}
    \caption{Certified accuracy as a function of the $\ell_2$ adversarial perturbation bound, 
    when varying levels of Gaussian noise $\sigma \in \{0.25, 0.5, 1.0\}$. 
    Bounds are computed with $100{,}000$ samples per run on CIFAR-10, and $10{,}000$ on ImageNet.}
    \label{fig:acc_vs_radius}
\end{figure}

On both CIFAR-10 and ImageNet we outperform the state-of-the-art denoised smoothing approaches (i.e.,~\citet{salman2020denoised} and \citet{leeprovable}) in every setting;
see Table~\ref{tab:imagenet_sota} and Table~\ref{tab:cifar10_sota}, as well as
Figure~\ref{fig:acc_vs_radius} for detailed results.
Perhaps even more impressively, we also outperform models trained with randomized smoothing at low $\varepsilon$ distortions {($\epsilon \leq 0.5$ on CIFAR-10, and $\epsilon\leq 2$ on ImageNet)}, and nearly match them at high $\varepsilon$. 
Even though these randomized smoothing techniques train their models end-to-end and
specifically design these models to have high accuracy on Gaussian noise,
we find that our approach's use of
off-the-shelf models yields superior robustness {(and much higher clean accuracy as an added bonus)}.

Interestingly, we find that using a diffusion model to perform the denoising step gives its most significant benefits when $\sigma$ and $\varepsilon$ are small: for example, while we reach $71.1\%$ top-1 accuracy at $\varepsilon=0.5$ on ImageNet, an improvement over prior work of $+14$ percentage points, when we reach $\varepsilon=3$ our scheme is $7$ percentage points
worse than state-of-the-art.
Our hypothesis for this effect, which we explore further in Section~\ref{sec:analysis}, is that diffusion models are prone to ``hallucinate'' content when denoising extremely noisy images. Thus, instead of reinforcing the signal from the correct class, the diffusion model generates a signal from another class, thereby fooling the classifier.

\paragraph{CIFAR-10 ablation.}
{The off-the-shelf classifiers we use were pretrained on larger datasets than respectively CIFAR-10 and ImageNet. It is well known that the use of additional data can boost robustness, both for empirical~\citep{schmidt2018adversarially} and certified~\citep{salman2019provably} defenses.
To investigate the role played by the pretrained model, we repeat our CIFAR-10 experiment using a standard 
Wide-ResNet-28-10 model~\citep{zagoruyko2016wide} trained solely on CIFAR-10 to 95.2\% accuracy. The results with this classifier (see Table~\ref{tab:cifar10_all_results:resnet}) outperform prior denoised smoothing approaches, and are competitive with prior randomized smoothing results up to $\epsilon=0.5$.}

{The ViT classifier outperforms the ResNet because it is more robust to the distribution shift introduced by the noising-and-denoising procedure. To alleviate this, we can further \emph{finetune} the classifier on denoised images $\texttt{denoise}(x+\delta)$ from the CIFAR-10 training set. This defense is thus not strictly ``off-the-shelf'' anymore (although finetuning is negligible compared to the training time of the diffusion model and classifier). Table~\ref{tab:cifar10_all_results:resnet_finetuned} shows that a finetuned Wide-ResNet achieves comparable-or-better results than a non-finetuned ViT. Thus, with a minimal amount of training, we also surpass prior randomized smoothing results \emph{without relying on any external data}. If we finetune the ViT model (Table~\ref{tab:cifar10_all_results:vit_finetuned}), we further improve our defense's clean accuracy and certified robustness at $\epsilon \leq 0.5$ by a couple of percentage points. Our ablation is summarized in Table~\ref{tab:cifar10_ours}.}

\begin{table}
    \vspace{-1em}
    \centering
        \resizebox{1.0\textwidth}{!}{
    \begin{tabular}{@{}l c c r r r r@{}}
        & & & \multicolumn{4}{c}{Certified Accuracy at $\varepsilon$ (\%)} \\
         \cmidrule{4-7}
        Method & Off-the-shelf & Extra data & $0.25$ & $0.5$ & $0.75$ & $1.0$ \\
        \toprule
        Wide-ResNet &  \CIRCLE & \xmark & $^{(83.8)}$70.6  & $^{(83.8)}$55.7  & $^{(83.8)}$40.0  & $^{(65.8)}$26.1\\
        ViT &  \CIRCLE & \cmark & $^{(88.1)}$76.7  & $^{(88.1)}$63.0  & $^{(88.1)}$45.3  & $^{(77.0)}$32.1\\
        Wide-ResNet +finetune & \LEFTcircle & \xmark & $^{(85.9)}$76.7  & \bf$^{(85.9)}$63.8  & \bf $^{(85.9)}$49.5  & \bf $^{(74.5)}$36.4\\
        ViT +finetune & \LEFTcircle & \cmark & \bf $^{(91.2)}$79.3  & \bf $^{(91.2)}$65.5  & \bf$^{(91.2)}$48.7  & \bf$^{(81.5)}$35.5\\
        \bottomrule
    \end{tabular}
    }
    \caption{Summary of our ablation on CIFAR-10. The diffusion model and Wide-ResNet classifier are trained solely on CIFAR-10, while the ViT classifier is pretrained on a larger dataset. The finetuning results are obtained by taking an off-the-shelf diffusion model and classifier, and tuning the classifier on noised-then-denoised images from CIFAR-10.}
    \label{tab:cifar10_ours}
\end{table}

\section{Analysis and Discussion}
\label{sec:analysis}

We achieve state-of-the-art certified accuracy using diffusion models
despite the fact that we are not using these models as \emph{diffusion models}
but rather \emph{trivial denoisers}.
That is, instead of leveraging the fact that diffusion models can \emph{iteratively} refine images across a \emph{range of noise levels}, we simply apply the diffusion model \emph{once} for a \emph{fixed} noise level, to perform one-shot denoising.  

In this section we study why this approach outperforms prior work that trained straightforward denoisiers for denoised smoothing~\citep{salman2020denoised}, and why using diffusion models for one-shot denoising performs better than the more involved iterative diffusion process. Last we show promising results of multi-step diffusion using an advanced deterministic sampler.

\subsection{Full diffusion versus one-shot denoising}\label{full_vs_one_shot}

When used as generative models, diffusion models perform denoising through an iterative process that repeatedly refines an estimate of the final denoised image. When given an image $x_t$ with noise of magnitude corresponding to some diffusion timestep $t$, the model first predicts a one-shot estimate of the denoised image $x_0$, and then constructs an estimate $x_{t-1}$ of the noised image at timestep $t-1$ by interpolating (with appropriate weights) between $x_0$, $x_t$ \emph{and fresh isotropic Gaussian noise} $\mathcal{N}(0, I)$. The diffusion process is then applied recursively at timestep $t-1$.

Intuitively, it may be expected that when using a diffusion model as a denoiser, one-shot denoising
will produce more faithful results than the full iterative reverse-diffusion process. Indeed, each step of the reverse-diffusion process \emph{destroys information} about the original image, since each step adds fresh Gaussian noise to the image. Thus, information theoretically at least, it should be easier to denoise an image in one-shot than over multiple iterations. 

\begin{figure}[h]
    \centering
        \includegraphics[width=1.0\textwidth]{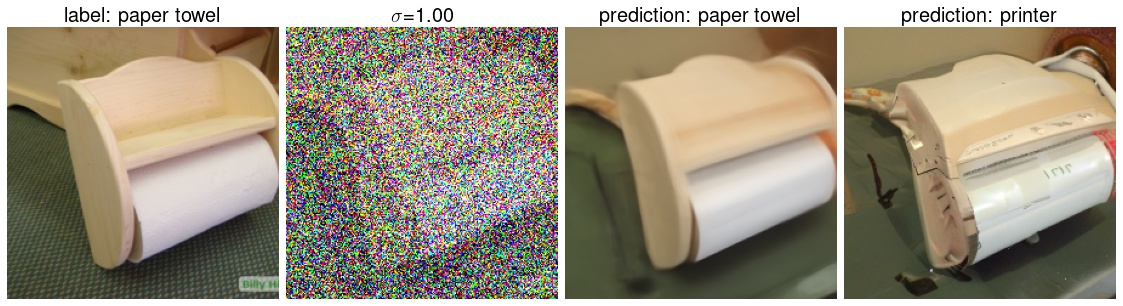}
    \caption{Intuitive examples for why multi-step denoised images are less recognized by the classifier. From left to right: clean images, noisy images with $\sigma=1.0$, one-step denoised images, multi-step denoised images. For the denoised images, we show the prediction by the pretrained BEiT model.} 
    \label{fig:counter_example}
\end{figure}

We find that this is indeed the case. While the full reverse-diffusion process produces denoised images with more finegrained details (which is a good property for generating photorealistic images from scratch), these details are often not actually faithful to the original image we want to denoise. Instead, diffusion models are prone to ``hallucinate'' salient detailed features during the iterative denoise-and-noise process.
We illustrate some examples of this hallucination phenomenon in Figure~\ref{fig:counter_example}. Here, we noise an original image (on the left) with large Gaussian noise ($\sigma=1$) and then apply either the full reverse-diffusion process (rightmost image) or a one-shot denoising at the appropriate timestep (2nd image to the right).
As we can see, one-shot denoising produces mostly faithful, but blurry, reconstructions of the original image, with finegrained details lost due to noise. In contrast, iterative denoising ``invents'' new details that result in images that are ultimately more photorealistic but semantically different from the starting image. Additional examples (with multiple random seeds) are in Figure~\ref{fig:comparison} and Figure~\ref{fig:counter_example_appendix} in the Appendix.


\subsection{Training on restricted noise levels}

Given that one-shot denoising performs better than full multi-shot denoising,
we now turn to understanding our next question: if we are just  using diffusion models as one-shot denoisers, then why do diffusion models perform better compared to the straightforward denoisers trained in prior work~\citep{salman2020denoised}?
To investigate this, we train seven new diffusion models on CIFAR-10 with varying levels of Gaussian noise---all the way towards a model trained on a single noise level, i.e., a straightforward denoiser.

Recall that during standard training of a diffusion model, we sample a timestep $T$
uniformly from some range, add noise according to this timestep, and then train the model to
predict the noise that has been added.
The only difference between this process and the standard denoised smoothing training process~\citep{salman2020denoised} is the fact that here we are training on multiple levels of Gaussian noise simultaneously.
Therefore we now perform a comparative analysis of models trained on more restrictive noise levels.
We select seven different levels of noise:

\begin{itemize}
    \item Three models are trained exclusively on Gaussian noise of fixed standard deviation of respectively $\sigma=0.25$, $\sigma=0.5$, or $\sigma=1.0$. This is identical to training a ``straightforward'' denoiser on noise of a fixed magnitude.
    \item One model is trained on all three noise levels at the same time.
    \item Two models are trained on noise uniformly selected from $\sigma \in [0,0.25]$, and $\sigma \in [0,1.0]$.
    \item One model is trained using the full range of noise, from $\sigma \in [0, S]$ for some $S \gg 1$ (the exact value of $S$ depends on the chosen noise schedule for the diffusion model).
\end{itemize}

We then evaluate the clean accuracy of an off-the-shelf ViT model on each image when denoised (in one shot) with each of these diffusion models, where the images are noised with a standard deviation of either $\sigma=0.25$, $\sigma=0.5$, or $\sigma=1.0$. The results are summarized in Table~\ref{tab:noise_levels}.

\begin{table}[h]
    \centering
    \begin{tabular}{@{} lrrr @{}}
    \toprule
    & \multicolumn{3}{c}{Noise at evaluation}\\
    \cmidrule{2-4}
    Training noise & $\sigma=0.25$ & $\sigma=0.5$ & $\sigma=1.0$ \\
    \midrule
         $\sigma \in \{0.25\}$ & 79.0 & 16.2 & 9.8 \\
    $\sigma \in \{0.5\}$ & 14.5 & 60.1 & 15.4 \\
    $\sigma \in \{1.0\}$ & 13.9 & 13.5 & 35.5 \\
    $\sigma \in \{0.25, 0.5, 1.0\}$ & 81.6 & 68.1 & 43.0 \\
    $\sigma \in [0, 0.25]$ & 84.5 & 14.5 & 9.9 \\
    $\sigma \in [0, 1.0]$ & 84.0 & 71.6 & \bf 46.0 \\
    $\sigma \in [0, S \gg 1]$ (standard) & \bf 85.5 & \bf  72.3 & 44.8 \\
\bottomrule
    \end{tabular}
    \caption{Clean accuracy of an off-the-shelf ViT classifier on images denoised with a diffusion model trained on restricted levels of Gaussian noise.
    Diffusion models trained on more diverse noise ranges yield higher accuracy on one-shot denoised images, even compared to models trained on the specific noise level used at evaluation time.}
    \label{tab:noise_levels}
\end{table}

As expected, training a new model on any one individual noise level, and then using that model to denoise images at that noise
level, gives high downstream accuracy: for example, training a diffusion model using $\sigma=0.25$ noise and then evaluating at this same noise level gives $79\%$ accuracy.
However if we then try and use this model to denoise images at a different noise level---say $\sigma=0.5$---the
accuracy of the classifier drops to just $16\%$.
If we train the diffusion model directly on $\sigma=0.5$ noise, we instead get a much better classification accuracy of $60.1\%$, but without good generalization to lower or higher noise levels. Similarly, training on noise of $\sigma=1.0$ only gives good results when denoising images with the same noise level.

More surprisingly, however, is that training on all three noise levels \emph{simultaneously} gives better accuracy for denoising images at each noise level, compared to a diffusion model trained specifically and solely for that noise level. For example, when denoising images with $\sigma=0.5$ Gaussian noise, we get a classification accuracy of $68.1\%$ when the diffusion model is trained on that noise level \emph{and} additional lower and higher noise levels---a value $8\%$ higher than the accuracy of $60.1\%$ we get when training the diffusion model solely on $\sigma=0.5$ noise.

If we train on more granular noise levels, either in $[0, 0.25]$ or in the full interval $[0, 1]$, the classification accuracy on denoised images at the three individual noise levels further increases by a few percentage points. Quite surprisingly, the standard training regime which trains the diffusion model on noise from a larger range $[0, S]$ for some $S \gg 1$ further improves the denoising capabilities at low noise levels ($\sigma=0.25$ and $\sigma=0.5$), but slightly harms the accuracy for larger noise ($\sigma=1.0$).

From this experiment, we can conclude that the (full) training process of diffusion models leads to much better, and more generalizable, one-shot denoising capabilities than when training a standalone denoiser on a single noise level as in prior work.

\subsection{Advanced deterministic multi-step sampler}
In section~\ref{full_vs_one_shot}, we found that the denoised images from full multi-step diffusion have a tendency to deviate from the original clean image. This could be due to the stochastic nature of the full reverse-diffusion process, since at each step a random noise is added. We notice a line of work~\citep{song2021ddim,karras2022edm} on fast deterministic sampling of diffusion models. We show that with such an advanced sampler, multi-step diffusion is able to beat one-shot denoising. 

We consider the deterministic EDM sampler proposed by \cite{karras2022edm}. We compare the recognizability of images denoised by EDM sampler and one-shot denoising. We adapt EDM sampler for image denoising by setting the maximum noise sigma of the sampling noise schedule to be the noise level found by \autoref{eq:solve_sigma}. We use the suggested sampler setting from \cite{karras2022edm} on CIFAR-10, where 18 reverse steps with 35 evaluations of the diffusion model are performed for each example. The result is summarized in Table~\ref{tab:edm_sampler}. We can see that the deterministic EDM sampler is superior over one-shot denoising.

\begin{table}[h]
    \centering
    \begin{tabular}{@{} llrrr @{}}
    \toprule
      Classifier & Method & $\sigma=0.25$ & $\sigma=0.5$ & $\sigma=1.0$ \\
    \midrule
    \multirow{2}{*}{Wide-ResNet} &   One-shot denoising & 81.3  & 64.0 & 35.8 \\
    & EDM sampler & 85.0 & 73.0  & 53.8 \\
    \midrule 
    \multirow{2}{*}{VIT} &   One-shot denoising & 84.9 & 71.6 & 50.8 \\
    & EDM sampler & 86.1 & 73.1 & 54.0 \\
\bottomrule
    \end{tabular}
    \caption{Clean accuracy (average over 5 runs) of off-the-shelf CIFAR-10 classifiers evaluated on images denoised by one-shot denoising and EDM sampler~\citep{karras2022edm}.}
    \label{tab:edm_sampler}
\end{table}

\section{Conclusion}

At present, training certified adversarially robust deep learning models requires specialized
techniques explicitly designed for the purpose of performing provably robust classification \citep{cohen2019certified}.
While this has proven effective, these models are extremely difficult to train to high accuracy, and degrade clean accuracy significantly.

We suggest an alternative approach is possible.
By exclusively making use of off-the-shelf models designed to be state-of-the-art at
classification and image denoising, we can leverage the vast resources dedicated to
training highly capable models for the new purpose of robust classification.

\bibliography{ref}
\bibliographystyle{iclr2023_conference}

\clearpage
\appendix
\section{Appendix}

\begin{table}[!htb]

    \begin{subfigure}[b]{0.48\textwidth}
    \centering
    \begin{tabular}{@{}lrrrrrr@{}}
         & \multicolumn{5}{c}{Certified Accuracy at $\varepsilon$ (\%)} \\
         \cmidrule{2-6}
        Noise & $0.0$ & $0.25$ & $0.5$ & $0.75$ & $1.0$\\
         \midrule
        $\sigma=0.25$ & 83.8 & 70.6 & 55.7 & 40.0 & 0.0 \\ 
        $\sigma=0.5$  & 65.8 & 54.7 & 43.7 & 34.2 & 26.1 \\ 
        $\sigma=1.0$  & 33.2 & 28.0 & 22.8 & 18.0 & 13.6 \\
        \bottomrule
    \end{tabular}
    \caption{Wide-ResNet}
    \label{tab:cifar10_all_results:resnet}
    \end{subfigure}
    \hfill
    \begin{subfigure}[b]{0.48\textwidth}
    \centering
    \begin{tabular}{@{}lrrrrrr@{}}
         & \multicolumn{5}{c}{Certified Accuracy at $\varepsilon$ (\%)} \\
         \cmidrule{2-6}
        Noise & $0.0$ & $0.25$ & $0.5$ & $0.75$ & $1.0$\\
         \midrule
        $\sigma=0.25$ & 85.9 & 76.7 & 63.8 & \bf 49.5 & 0.0 \\
        $\sigma=0.5$  & 74.5 & 66.0 & 56.1 & 45.7 & \bf 36.4 \\
        $\sigma=1.0$  & 55.1 & 48.7 & 42.3 & 35.8 & 29.9 \\
        \bottomrule
    \end{tabular}
    \caption{Finetuned Wide-ResNet}
    \label{tab:cifar10_all_results:resnet_finetuned}
    \end{subfigure}
    \\
    \\
    \begin{subfigure}[b]{0.48\textwidth}
    \centering
    \begin{tabular}{@{}lrrrrrr@{}}
         & \multicolumn{5}{c}{Certified Accuracy at $\varepsilon$ (\%)} \\
         \cmidrule{2-6}
        Noise & $0.0$ & $0.25$ & $0.5$ & $0.75$ & $1.0$\\
         \midrule
        $\sigma=0.25$ & 88.1 & 76.7 & 63.0 & 45.3 & 0.0 \\
        $\sigma=0.5$  & 77.0 & 65.8 & 53.4 & 41.8 & 32.1 \\
        $\sigma=1.0$  & 49.5 & 40.3 & 33.3 & 26.1 & 20.2 \\
        \bottomrule
    \end{tabular}
    \caption{ViT}
    \label{tab:cifar10_all_results:vit}
    \end{subfigure}
    \hfill
    \begin{subfigure}[b]{0.48\textwidth}
    \centering
    \begin{tabular}{@{}lrrrrrr@{}}
         & \multicolumn{5}{c}{Certified Accuracy at $\varepsilon$ (\%)} \\
         \cmidrule{2-6}
        Noise & $0.0$ & $0.25$ & $0.5$ & $0.75$ & $1.0$\\
         \midrule
        $\sigma=0.25$ & \bf 91.2 & \bf 79.3 & \bf 65.5 & 48.7 & 0.0 \\
        $\sigma=0.5$ & 81.5 & 67.0 & 56.1 & 45.3 & 35.5 \\
        $\sigma=1.0$ & 65.1 & 48.4 & 41.7 & 35.2 & 29.0 \\
        \bottomrule
    \end{tabular}
    \caption{Finetuned ViT}
    \label{tab:cifar10_all_results:vit_finetuned}
    \end{subfigure}
    
    \caption{Certified accuracy of four different classifiers on CIFAR-10 at varying levels of Gaussian noise $\sigma$, all using the same diffusion model.}
    \label{tab:cifar10_all_results}
\end{table}

\begin{table}[!htb]
    \centering
    \vspace{-1em}
    \begin{tabular}{@{}lrrrrrrr@{}}
         & \multicolumn{6}{c}{Certified Accuracy at $\varepsilon$ (\%)} \\
         \cmidrule{2-7}
        Noise & $0.0$ & $0.5$ & $1.0$ & $1.5$ & $2.0$ & $3.0$\\
         \midrule
        $\sigma=0.25$ & \bf 82.8 & \bf 71.1 & 0.0 & 0.0 & 0.0 & 0.0 \\
        $\sigma=0.5$ & 77.1 & 67.8 & \bf 54.3 & \bf 38.1 & 0.0 & 0.0 \\
        $\sigma=1.0$ & 60.0 & 50.0 & 42.0 & 35.5 & \bf 29.5 & \bf 13.1 \\
        \bottomrule
    \end{tabular}

    \caption{Certified accuracy on ImageNet for varying levels of Gaussian noise $\sigma$.}
    \label{tab:imagenet_all_results}
\end{table}

\begin{figure}[h]
    \vspace{-1em}
    \begin{subfigure}[b]{0.95\textwidth}
     \centering
        \includegraphics[width=0.75\textwidth]{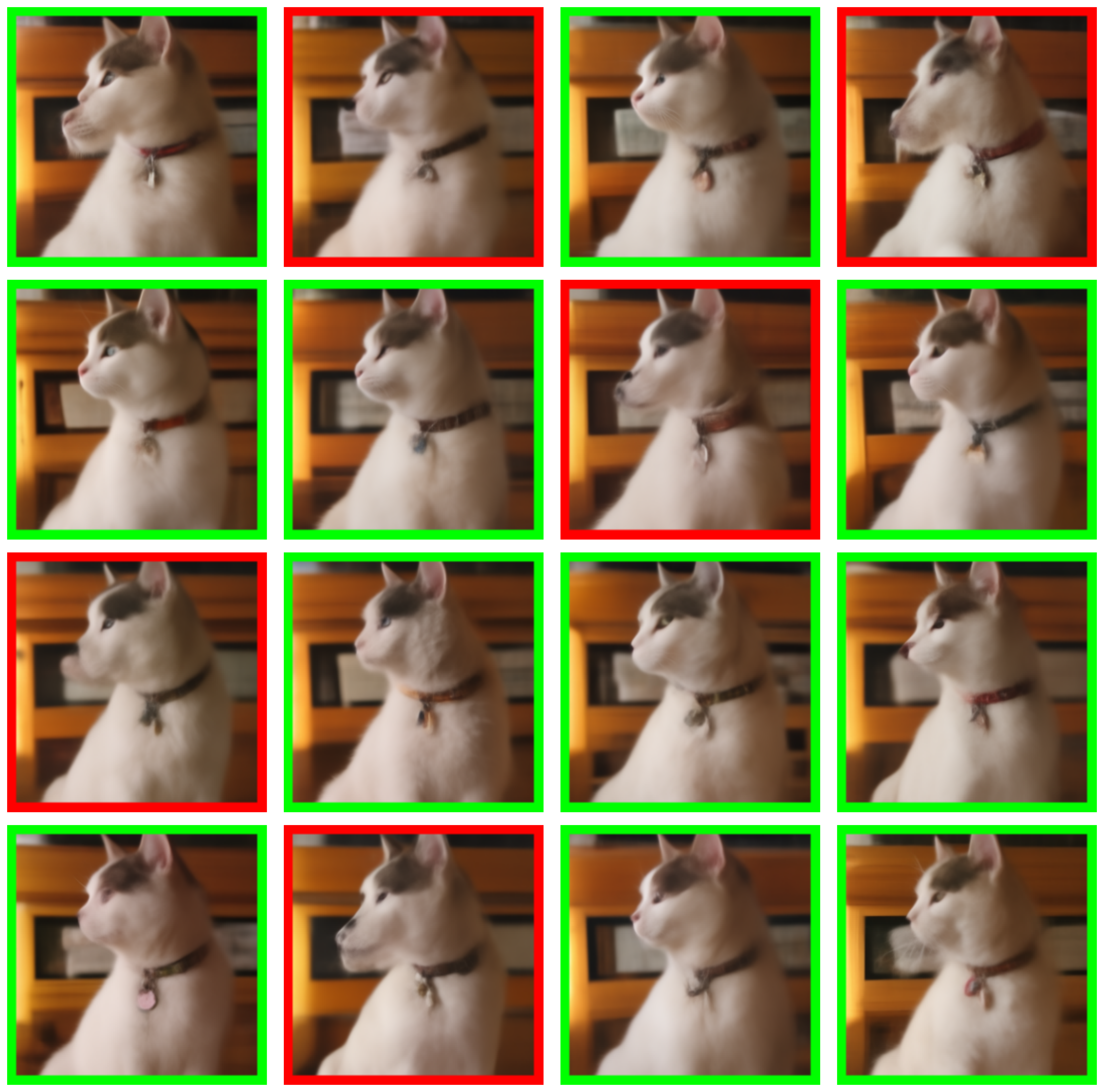}
    \caption{One-shot denoised images ($\sigma=1.00$)}
    \end{subfigure}
    \begin{subfigure}[b]{0.95\textwidth}
     \centering
        \includegraphics[width=0.75\textwidth]{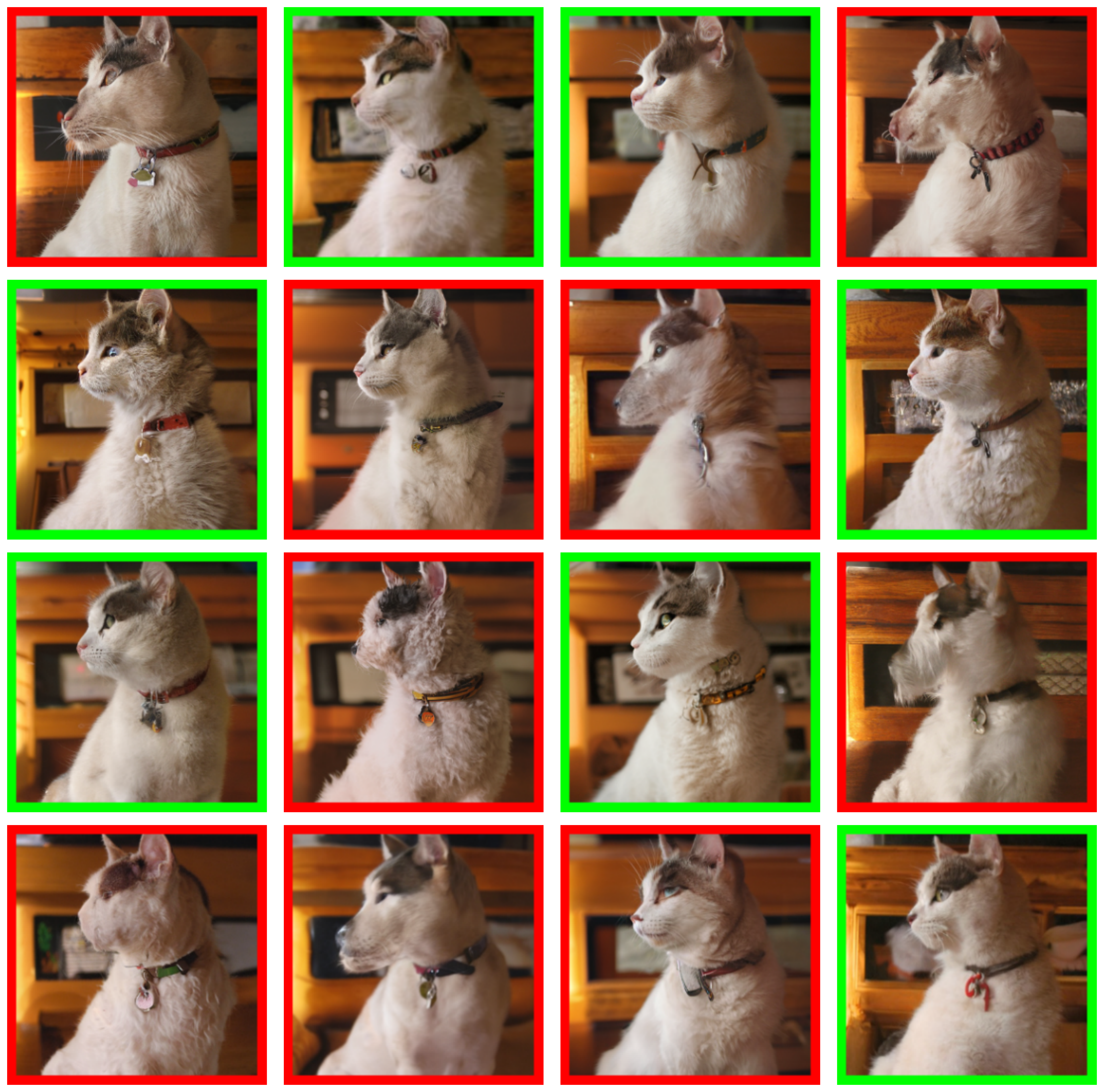}
    \caption{Multi-step denoised images ($\sigma=1.00$)} 
    \end{subfigure}
    \caption{Qualitative comparison of one-shot denoising and multi-step denoising. We show denoised images under random Gaussian noise ($\sigma=1.00$). A green border is applied when the denoised images are correctly classified while a red border means that the classifier misclassifies the image.}
    \label{fig:comparison}
    \vspace{-3em}
\end{figure}

\begin{figure}[h]
    \centering
        \includegraphics[width=1.0\textwidth]{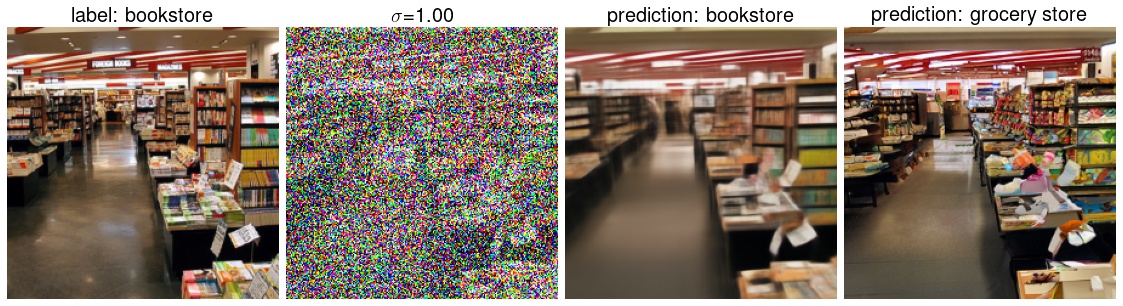}
        \includegraphics[width=1.0\textwidth]{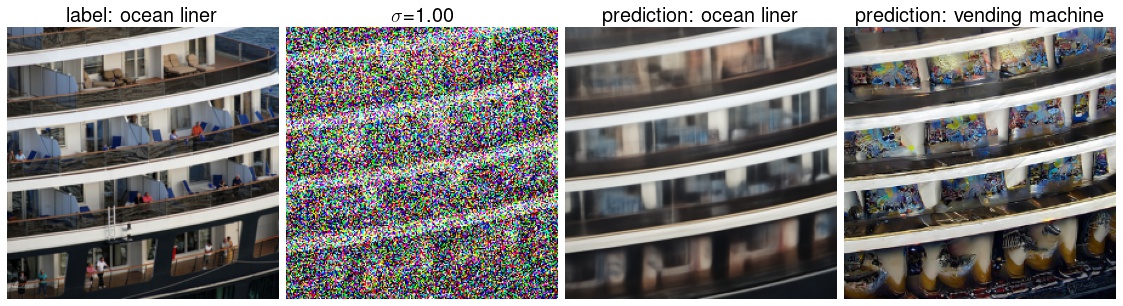}
    \caption{Additional intuitive examples for why multi-step denoised images are less recognized by the classifier. From left to right: clean images, noisy images with $\sigma=1.0$, one-step denoised images, multi-step denoised images. For the denoised images, we show the prediction by the pretrained BEiT model.} 
    \label{fig:counter_example_appendix}
\end{figure}

\end{document}

%% file: math_commands.tex
\usepackage{amsmath,amsfonts,bm}

\def\eqref#1{equation~\ref{#1}}
\def\Eqref#1{Equation~\ref{#1}}

\def\1{\bm{1}}

\DeclareMathAlphabet{\mathsfit}{\encodingdefault}{\sfdefault}{m}{sl}
\SetMathAlphabet{\mathsfit}{bold}{\encodingdefault}{\sfdefault}{bx}{n}













%% file: iclr2023_conference.bbl
\begin{thebibliography}{31}
\providecommand{\natexlab}[1]{#1}
\providecommand{\url}[1]{\texttt{#1}}
\expandafter\ifx\csname urlstyle\endcsname\relax
  \providecommand{\doi}[1]{doi: #1}\else
  \providecommand{\doi}{doi: \begingroup \urlstyle{rm}\Url}\fi

\bibitem[Athalye et~al.(2018)Athalye, Carlini, and
  Wagner]{athalye2018obfuscated}
Anish Athalye, Nicholas Carlini, and David Wagner.
\newblock Obfuscated gradients give a false sense of security: Circumventing
  defenses to adversarial examples.
\newblock In \emph{International Conference on Machine Learning}, pp.\
  274--283. PMLR, 2018.

\bibitem[Bao et~al.(2022)Bao, Dong, and Wei]{bao2021beit}
Hangbo Bao, Li~Dong, and Furu Wei.
\newblock {BEiT}: {BERT} pre-training of image transformers.
\newblock In \emph{International Conference on Learning Representations}, 2022.

\bibitem[Biggio et~al.(2013)Biggio, Corona, Maiorca, Nelson, {\v{S}}rndi{\'c},
  Laskov, Giacinto, and Roli]{biggio2013evasion}
Battista Biggio, Igino Corona, Davide Maiorca, Blaine Nelson, Nedim
  {\v{S}}rndi{\'c}, Pavel Laskov, Giorgio Giacinto, and Fabio Roli.
\newblock Evasion attacks against machine learning at test time.
\newblock In \emph{Joint European conference on machine learning and knowledge
  discovery in databases}, pp.\  387--402. Springer, 2013.

\bibitem[Carlini \& Wagner(2017)Carlini and Wagner]{carlini2017towards}
Nicholas Carlini and David Wagner.
\newblock Towards evaluating the robustness of neural networks.
\newblock In \emph{2017 IEEE symposium on security and privacy}, pp.\  39--57.
  IEEE, 2017.

\bibitem[Cohen et~al.(2019)Cohen, Rosenfeld, and Kolter]{cohen2019certified}
Jeremy Cohen, Elan Rosenfeld, and Zico Kolter.
\newblock Certified adversarial robustness via randomized smoothing.
\newblock In \emph{International Conference on Machine Learning}, pp.\
  1310--1320. PMLR, 2019.

\bibitem[Deng et~al.(2009)Deng, Dong, Socher, Li, Li, and
  Fei-Fei]{deng2009imagenet}
Jia Deng, Wei Dong, Richard Socher, Li-Jia Li, Kai Li, and Li~Fei-Fei.
\newblock {ImageNet}: A large-scale hierarchical image database.
\newblock In \emph{2009 IEEE conference on computer vision and pattern
  recognition}, pp.\  248--255. Ieee, 2009.

\bibitem[Dhariwal \& Nichol(2021)Dhariwal and Nichol]{dhariwal2021diffusion}
Prafulla Dhariwal and Alexander Nichol.
\newblock Diffusion models beat {GANs} on image synthesis.
\newblock \emph{Advances in Neural Information Processing Systems}, 34, 2021.

\bibitem[Dosovitskiy et~al.(2021)Dosovitskiy, Beyer, Kolesnikov, Weissenborn,
  Zhai, Unterthiner, Dehghani, Minderer, Heigold, Gelly,
  et~al.]{dosovitskiy2020image}
Alexey Dosovitskiy, Lucas Beyer, Alexander Kolesnikov, Dirk Weissenborn,
  Xiaohua Zhai, Thomas Unterthiner, Mostafa Dehghani, Matthias Minderer, Georg
  Heigold, Sylvain Gelly, et~al.
\newblock An image is worth 16x16 words: Transformers for image recognition at
  scale.
\newblock In \emph{International Conference on Learning Representations}, 2021.

\bibitem[Gowal et~al.(2018)Gowal, Dvijotham, Stanforth, Bunel, Qin, Uesato,
  Arandjelovic, Mann, and Kohli]{gowal2018effectiveness}
Sven Gowal, Krishnamurthy Dvijotham, Robert Stanforth, Rudy Bunel, Chongli Qin,
  Jonathan Uesato, Relja Arandjelovic, Timothy Mann, and Pushmeet Kohli.
\newblock On the effectiveness of interval bound propagation for training
  verifiably robust models.
\newblock \emph{arXiv preprint arXiv:1810.12715}, 2018.

\bibitem[Ho et~al.(2020)Ho, Jain, and Abbeel]{ho2020denoising}
Jonathan Ho, Ajay Jain, and Pieter Abbeel.
\newblock Denoising diffusion probabilistic models.
\newblock \emph{Advances in Neural Information Processing Systems},
  33:\penalty0 6840--6851, 2020.

\bibitem[Horv{\'a}th et~al.(2022{\natexlab{a}})Horv{\'a}th, M{\"u}ller,
  Fischer, and Vechev]{horvath2021boosting}
Mikl{\'o}s~Z Horv{\'a}th, Mark~Niklas M{\"u}ller, Marc Fischer, and Martin
  Vechev.
\newblock Boosting randomized smoothing with variance reduced classifiers.
\newblock In \emph{International Conference on Learning Representations},
  2022{\natexlab{a}}.

\bibitem[Horv{\'a}th et~al.(2022{\natexlab{b}})Horv{\'a}th, M{\"u}ller,
  Fischer, and Vechev]{horvath2022robust}
Mikl{\'o}s~Z Horv{\'a}th, Mark~Niklas M{\"u}ller, Marc Fischer, and Martin
  Vechev.
\newblock Robust and accurate--compositional architectures for randomized
  smoothing.
\newblock \emph{arXiv preprint arXiv:2204.00487}, 2022{\natexlab{b}}.

\bibitem[Jeong \& Shin(2020)Jeong and Shin]{jeong2020consistency}
Jongheon Jeong and Jinwoo Shin.
\newblock Consistency regularization for certified robustness of smoothed
  classifiers.
\newblock \emph{Advances in Neural Information Processing Systems},
  33:\penalty0 10558--10570, 2020.

\bibitem[Jeong et~al.(2021)Jeong, Park, Kim, Lee, Kim, and
  Shin]{jeong2021smoothmix}
Jongheon Jeong, Sejun Park, Minkyu Kim, Heung-Chang Lee, Do-Guk Kim, and Jinwoo
  Shin.
\newblock Smoothmix: Training confidence-calibrated smoothed classifiers for
  certified robustness.
\newblock \emph{Advances in Neural Information Processing Systems},
  34:\penalty0 30153--30168, 2021.

\bibitem[Karras et~al.(2022)Karras, Aittala, Aila, and Laine]{karras2022edm}
Tero Karras, Miika Aittala, Timo Aila, and Samuli Laine.
\newblock Elucidating the design space of diffusion-based generative models.
\newblock \emph{Advances in Neural Information Processing Systems}, 2022.

\bibitem[Lecuyer et~al.(2019)Lecuyer, Atlidakis, Geambasu, Hsu, and
  Jana]{lecuyer2019certified}
Mathias Lecuyer, Vaggelis Atlidakis, Roxana Geambasu, Daniel Hsu, and Suman
  Jana.
\newblock Certified robustness to adversarial examples with differential
  privacy.
\newblock In \emph{2019 IEEE Symposium on Security and Privacy}, pp.\
  656--672. IEEE, 2019.

\bibitem[Lee(2021)]{leeprovable}
Kyungmin Lee.
\newblock Provable defense by denoised smoothing with learned score function.
\newblock \emph{ICLR Workshop on Security and Safety in Machine Learning
  Systems}, 2021.

\bibitem[Mirman et~al.(2018)Mirman, Gehr, and Vechev]{mirman2018differentiable}
Matthew Mirman, Timon Gehr, and Martin Vechev.
\newblock Differentiable abstract interpretation for provably robust neural
  networks.
\newblock In \emph{International Conference on Machine Learning}, pp.\
  3578--3586. PMLR, 2018.

\bibitem[Nichol \& Dhariwal(2021)Nichol and Dhariwal]{nichol2021improved}
Alexander~Quinn Nichol and Prafulla Dhariwal.
\newblock Improved denoising diffusion probabilistic models.
\newblock In \emph{International Conference on Machine Learning}, pp.\
  8162--8171. PMLR, 2021.

\bibitem[Nie et~al.(2022)Nie, Guo, Huang, Xiao, Vahdat, and
  Anandkumar]{nie2022diffusion}
Weili Nie, Brandon Guo, Yujia Huang, Chaowei Xiao, Arash Vahdat, and Anima
  Anandkumar.
\newblock Diffusion models for adversarial purification.
\newblock \emph{arXiv preprint arXiv:2205.07460}, 2022.

\bibitem[Salman et~al.(2019)Salman, Li, Razenshteyn, Zhang, Zhang, Bubeck, and
  Yang]{salman2019provably}
Hadi Salman, Jerry Li, Ilya Razenshteyn, Pengchuan Zhang, Huan Zhang, Sebastien
  Bubeck, and Greg Yang.
\newblock Provably robust deep learning via adversarially trained smoothed
  classifiers.
\newblock \emph{Advances in Neural Information Processing Systems}, 32, 2019.

\bibitem[Salman et~al.(2020)Salman, Sun, Yang, Kapoor, and
  Kolter]{salman2020denoised}
Hadi Salman, Mingjie Sun, Greg Yang, Ashish Kapoor, and J~Zico Kolter.
\newblock Denoised smoothing: A provable defense for pretrained classifiers.
\newblock \emph{Advances in Neural Information Processing Systems},
  33:\penalty0 21945--21957, 2020.

\bibitem[Schmidt et~al.(2018)Schmidt, Santurkar, Tsipras, Talwar, and
  Madry]{schmidt2018adversarially}
Ludwig Schmidt, Shibani Santurkar, Dimitris Tsipras, Kunal Talwar, and
  Aleksander Madry.
\newblock Adversarially robust generalization requires more data.
\newblock \emph{Advances in Neural Information Processing Systems}, 31, 2018.

\bibitem[Sohl-Dickstein et~al.(2015)Sohl-Dickstein, Weiss, Maheswaranathan, and
  Ganguli]{sohl2015deep}
Jascha Sohl-Dickstein, Eric Weiss, Niru Maheswaranathan, and Surya Ganguli.
\newblock Deep unsupervised learning using nonequilibrium thermodynamics.
\newblock In \emph{International Conference on Machine Learning}, pp.\
  2256--2265. PMLR, 2015.

\bibitem[Song et~al.(2021)Song, Meng, and Ermon]{song2021ddim}
Jiaming Song, Chenlin Meng, and Stefano Ermon.
\newblock Denoising diffusion implicit models.
\newblock In \emph{International Conference on Learning Representations}, 2021.

\bibitem[Szegedy et~al.(2014)Szegedy, Zaremba, Sutskever, Bruna, Erhan,
  Goodfellow, and Fergus]{szegedy2013intriguing}
Christian Szegedy, Wojciech Zaremba, Ilya Sutskever, Joan Bruna, Dumitru Erhan,
  Ian Goodfellow, and Rob Fergus.
\newblock Intriguing properties of neural networks.
\newblock In \emph{International Conference on Learning Representations}, 2014.

\bibitem[Uesato et~al.(2018)Uesato, O’donoghue, Kohli, and
  Oord]{uesato2018adversarial}
Jonathan Uesato, Brendan O’donoghue, Pushmeet Kohli, and Aaron Oord.
\newblock Adversarial risk and the dangers of evaluating against weak attacks.
\newblock In \emph{International Conference on Machine Learning}, pp.\
  5025--5034. PMLR, 2018.

\bibitem[Wightman(2019)]{rw2019timm}
Ross Wightman.
\newblock Pytorch image models.
\newblock \url{https://github.com/rwightman/pytorch-image-models}, 2019.

\bibitem[Yang et~al.(2021)Yang, Li, Xu, Kailkhura, Xie, and
  Li]{yang2021certified}
Zhuolin Yang, Linyi Li, Xiaojun Xu, Bhavya Kailkhura, Tao Xie, and Bo~Li.
\newblock On the certified robustness for ensemble models and beyond.
\newblock \emph{arXiv preprint arXiv:2107.10873}, 2021.

\bibitem[Zagoruyko \& Komodakis(2016)Zagoruyko and
  Komodakis]{zagoruyko2016wide}
Sergey Zagoruyko and Nikos Komodakis.
\newblock Wide residual networks.
\newblock In \emph{British Machine Vision Conference}, 2016.

\bibitem[Zhai et~al.(2020)Zhai, Dan, He, Zhang, Gong, Ravikumar, Hsieh, and
  Wang]{zhai2020macer}
Runtian Zhai, Chen Dan, Di~He, Huan Zhang, Boqing Gong, Pradeep Ravikumar,
  Cho-Jui Hsieh, and Liwei Wang.
\newblock {MACER}: Attack-free and scalable robust training via maximizing
  certified radius.
\newblock In \emph{International Conference on Learning Representations}, 2020.

\end{thebibliography}
